\PassOptionsToPackage{dvipsnames}{xcolor}

\documentclass{hld2023}
\usepackage{booktabs}
\usepackage{bm}
\usepackage[shortlabels]{enumitem}

\definecolor{pltBlue}{HTML}{1f77b4}
\definecolor{pltOrange}{HTML}{ff7f0e}
\definecolor{pltGreen}{HTML}{2ca02c}
\definecolor{pltCornflowerBlue}{HTML}{6495ed}

\title[Predictive Sparse Manifold Transform]{Predictive Sparse Manifold Transform}

\hldauthor{
\Name{Yujia Xie}\thanks{Equal contribution.} \Email{yxie@gatech.edu}\\
\addr {Georgia Institute of Technology, Atlanta, USA}
\AND
\Name{Xinhui Li}\footnotemark[1] \Email{xinhuili@gatech.edu}\\
\addr {Georgia Institute of Technology, Atlanta, USA} 
\AND
\Name{Vince D. Calhoun} \Email{vcalhoun@gatech.edu}\\
\addr {Georgia Institute of Technology, Atlanta, USA} 
 }

\begin{document}

\maketitle

\begin{abstract}
We present \emph{Predictive Sparse Manifold Transform (PSMT)}, a minimalistic, interpretable and biologically plausible framework for learning and predicting natural dynamics. 
PSMT incorporates two layers where the first \emph{sparse coding} layer represents the input sequence as sparse coefficients over an overcomplete dictionary and the second \emph{manifold learning} layer learns a geometric embedding space that captures topological similarity and dynamic temporal linearity in sparse coefficients. 
We apply PSMT on a natural video dataset and evaluate the reconstruction performance with respect to contextual variability, the number of sparse coding basis functions and training samples. 
We then interpret the dynamic topological organization in the embedding space.
We next utilize PSMT to predict future frames compared with two baseline methods with a static embedding space. 
We demonstrate that PSMT with a dynamic embedding space can achieve better prediction performance compared to static baselines. 
Our work establishes that PSMT is an efficient unsupervised generative framework for prediction of future visual stimuli. 
\end{abstract}


\section{Introduction}

\textit{``Intelligence is the capacity of the brain to predict the future by analogy to the past." -- Jeff Hawkins} \cite{hawkins2004intelligence}

One fundamental problem of intelligence is how to precisely and efficiently predict future sensory stimuli based on prior knowledge and experience. 
Our perceived world can be usually viewed as a temporal sequence of natural scenes with smooth variability from one to the next. 
A sequence of natural images can be represented via sparse coefficients over an overcomplete dictionary, which characterizes spatially localized, oriented, bandpass receptive fields of simple cells in mammalian primary visual cortex \cite{olshausen1996emergence,olshausen1997sparse}. 
However, the learned sparse codes are not organized to reveal the topological structure of the spatiotemporal dynamics of visual stimuli. 
Sparse manifold transform (SMT) \cite{chen2018sparse} adds another \emph{manifold learning} layer \cite{vladymyrov2013locally} on top of the \emph{sparse coding} layer \cite{olshausen1996emergence} to capture topological similarity of the sparse coefficients. 
It models the continuous temporal transformation of the input sequence via a linear trajectory in the geometric embedding space.
Reconstruction of the input sequence from the geometric embedding space can well approximate the original sequence. 

Can we learn the structured spatiotemporal dynamics from the existing sequence and predict the future sequence leveraging the dynamic embedding space? 
In this study, we propose \emph{Predictive Sparse Manifold Transform (PSMT)}, a framework inspired by SMT and dynamic predictive coding \cite{chen2018sparse,jiang2022dynamic} to learn and predict spatiotemporal dynamics in natural scenes. 
We process a natural video as a sequence of image frames and apply PSMT on the dataset using different hyperparameters.
We interpret the topological organization of the geometric embedding matrix over time and demonstrate the dynamic nature of the embedding space. 
We next assess the reconstruction error with respect to the contextual variability in natural scenes.
We further predict the future frames using PSMT and two other baseline approaches with a static embedding space in two different contextual settings, and evaluate the prediction performance.
Our results demonstrate that using only two layers, PSMT with a dynamic embedding space can achieve better prediction performance compared to two static baselines, thus establishing a minimalistic, interpretable and biologically plausible framework for learning and predicting natural dynamics.

\section{Methods}
\subsection{Predictive Sparse Manifold Transform}
\label{psmt}

We outline the Predictive Sparse Manifold Transform (PSMT) framework here and expand its details in Appendix \ref{PSMT_detail}. Past temporal natural image sequence is processed as $\mathbf{X}=\{\bm{x}_t\}_{t=T-H+1}^{t=T-1}$ with $\bm{x}_t\in \mathbb{R}^N$, where $T$ is the index of current frame and $H$ is the number of previous frames. Future inputs $\mathbf{X^*}=\{\bm{x}_t\}_{t=T}^{t=T+K-1}$, where $K$ is the number of future frames to be predicted, are processed similarly and incorporated in PSMT iteratively as delayed inputs (Step 1).

In PSMT, we aim to predict future inputs $\bm{x}_{t=T-1+k}$ for $k=1,...,K$ given all previous inputs at $\bm{x}_{t=T-1+k-1}$. For the ease of presentation, we use footnotes \textit{prev} to represent $t=T-1+k-1$, \textit{current} to represent $t=T-1+k$, and \textit{future} to represent $t=T+k$.
\newcommand\mycommfont[1]{\scriptsize\ttfamily\textcolor{blue}{#1}}
\SetCommentSty{mycommfont}
\begin{algorithm2e}\label{PSMT_alg}
 \caption{Predictive Sparse Manifold Transform}
 \SetAlgoLined
  \KwIn{$T, H, K, M, N, f, \mathbf{X}=\{\bm{x}_t\}_{t=T-H+1}^{t=T-1}, \mathbf{X^*}=\{\bm{x}_t\}_{t=T}^{t=T+K-1}$\tcp*{$\mathbf{X^*}$: delayed input}} 
  \KwOut{$\hat{\bm{x}}_{t=T+k}$ for $k={1, ..., K}$}
  Sparse coding: $(\mathbf{\Phi}_M, \mathbf{A}) \leftarrow (\mathbf{\Phi}_M, \mathbf{A}:\mathbf{X}\approx \mathbf{\Phi}_M \mathbf{A}$) \tcp*{$\mathbf{\Phi}_M$: dictionary with $M$ items}
  \For{k = 1, ..., K}{
    Step 1: $\bm{x}_\text{current}:=\bm{x}_{t=T-1+k} \leftarrow \mathbf{X^*}_k$  \tcp*{receive most-recent input at $t=T-1+k$}
    Step 2: $\bm{\alpha}_\text{current}:=\bm{\alpha}_{t=T-1+k} \quad \text{s.t.} ~~ \bm{x}_{T-1+k}\approx \mathbf{\Phi}_M\bm{\alpha}_{T-1+k}$  \tcp*{get most-recent sparse code from $\mathbf{\Phi}_M$}
    Step 3: $\mathbf{A} \leftarrow [\mathbf{A}; \bm{\alpha}_\text{current}]$ \tcp*{update $\mathbf{A}$ to include most-recent sparse code}
    Step 4: $\mathbf{A^*}\ \leftarrow \mathbf{A}_{[t=T-H+k-1:t=T+k-1]}$ \tcp*{include most recent $H$ entries in $A$}
    Step 5: $\mathbf{P^*} \leftarrow \text{argmin}_{\mathbf{P}} \{ \| \mathbf{P} \mathbf{A^*}\mathbf{D}\|_F^2: \mathbf{P}\mathbf{V}\mathbf{P}^\top=\mathbf{I}$\}\tcp*{compute embedding matrix $\mathbf{P^*}$}
    Step 6: $\bm{\beta}^* \leftarrow \mathbf{P}^*\bm{\alpha}^*_\text{future}\approx 2\mathbf{P}^*\bm{\alpha}_\text{current} - \mathbf{P}^*\bm{\alpha}_\text{prev}$ \tcp*{predict next embedding at $t=T+k$}
    Step 7: $\bm{\alpha}^{\text{REC}}_\text{future}\leftarrow\text{argmin}_{\bm{\alpha}}  \{ \| \mathbf{P}^*\bm{\alpha}-\bm{\beta}^*\|_F^2 + \lambda \mathbf{z}^\top \bm{\alpha} : \bm{\alpha} \succeq 0\}$ \tcp*{reconstruct next sparse code}
    Step 8: $\hat{\bm{x}}_\text{future}=\mathbf{\Phi}_M \bm{\alpha}^{\text{REC}}_\text{future}$ \tcp*{estimate 1-step future input}
  }
\end{algorithm2e}

\begin{figure}[tbp]
\centerline{\includegraphics[width=15.5cm]{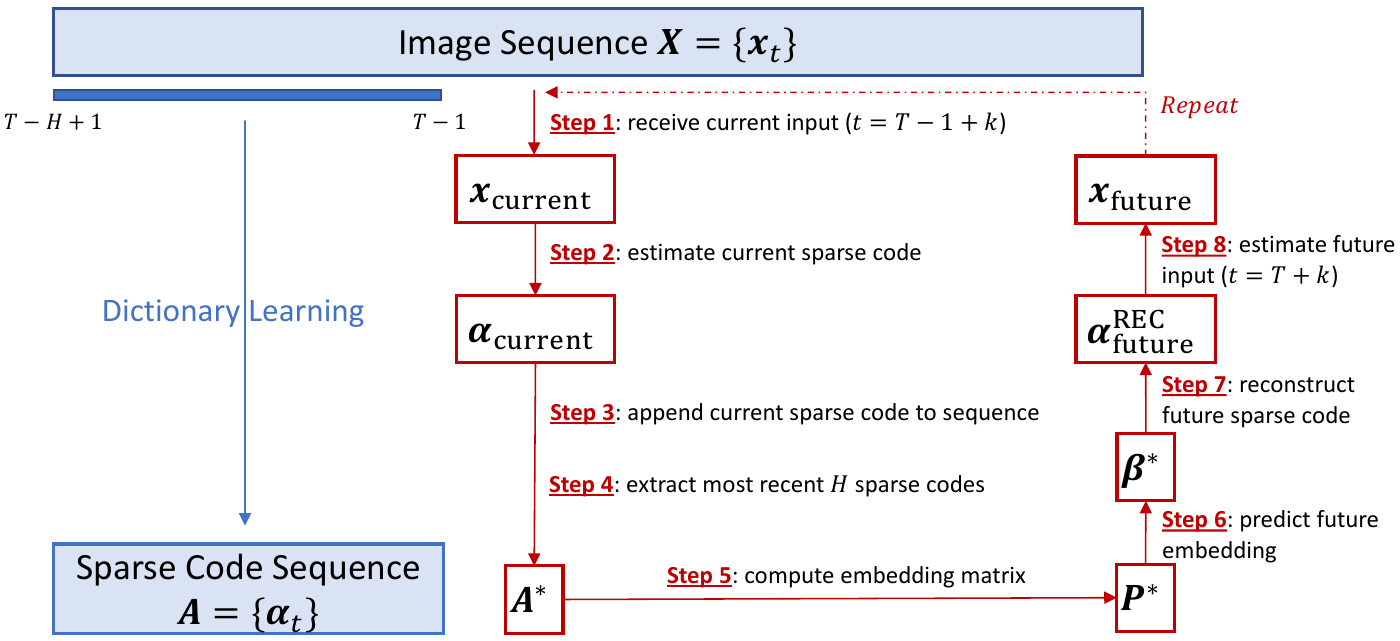}}
\caption{\textbf{PSMT algorithm illustration.}}
\label{fig_algo}
\end{figure}
\subsection{Experiment}

\textbf{Dataset and Preprocessing}~
We record the film \emph{Russian Ark} \cite{RussianArk} which was shot in one take and convert the video to a sequence of $10000$ image frames. 
Each frame includes $800 \times 800$ pixels, which is then downsampled to $200 \times 200$ pixels. 
We next extract a patch of $16 \times 16$ pixels at the center of the downsampled frame and perform standardization on each patch. 

\noindent \textbf{Hyperparameter Search}~
To investigate the optimal hyperparameters, we test $9$ combinations of hyperparameters including the number of sparse coding basis functions $M$ and the number of training frames $H$. 
We set $M=512, 1024, 2048$ to learn a $2\times$, $4\times$, $8\times$ overcomplete dictionary, respectively. For each $M$, we evaluate a training size of $2\times$, $2.5\times$, $3\times M$, each with a $\frac{3}{4}$ overlapping window. 
We compute the reconstruction mean squared error (MSE) for each combination. 

\noindent \textbf{Dynamic Topological Analysis}~
We analyze the topology of $\mathbf{\Phi}_M$ by looking at the embedding matrix $\mathbf{P^*}$ dynamically for each training frame $H$. Each column in $\mathbf{P^*}$ corresponds to each column in $\mathbf{\Phi}_M$ (i.e., each dictionary item). We compute the pairwise cosine-similarity among columns in $\mathbf{P^*}$ and construct a graph $\mathcal{G}$ with each node representing each dictionary item, and each edge representing whether the cosine-similarity between the nodes is greater than a threshold ($0.7$). Using the optimal hyperparameter combination from the hyperparameter search above, we have $\mathbf{P^*}_t$ centered at different time $t$ and consequently $\mathcal{G}_t$. We select a subset of nodes in $\mathcal{G}$ to visualize across time and analyze their connectivity behavior (i.e., neighbor count). We further investigate the implication of similarity in $\mathbf{P^*}$ to topological similarity between dictionary items.

\noindent \textbf{Reconstruction}~
Unlike SMT using all frames to reconstruct, PSMT relies only on $H$ frames prior to time $T$ to make future predictions. 
By analyzing the reconstruction performance with respect to $T$ and $H$, we can develop insights into tuning these two parameters for prediction. 
We implement the original SMT with all frames included and measure the reconstruction MSE at each frame.

\noindent \textbf{Prediction}~
We evaluate the prediction performance with respect to two hyperparameters -- start time $T$ and future prediction sequence length $K$ -- estimated from reconstruction result. 
We compare PSMT with two baseline methods. 
Baseline $1$ uses a static embedding matrix $\mathbf{P}^*$ and a static sparse coefficient $\bm{\alpha}^*$ obtained from $H$ training frames for all future predictions. 
Baseline $2$ also uses the same static $\mathbf{P}^*$ but updates the sparse coefficient $\bm{\alpha}_t$ across time. 
In contrast, PSMT utilizes a dynamic embedding matrix $\mathbf{P}^*$ as well as a dynamic sparse coefficient $\bm{\alpha}_t$, which is updated at each future frame $k$. 
We measure the log MSE between the reconstructed image patch from the sparse coefficient $\bm{\alpha}$ learned directly from inputs and the predicted image patch from Step 8 in PSMT (see Algorithm \ref{PSMT_alg}).

\begin{table}[tpb]
  \caption{\textbf{Hyperparameter search.} The mean $\pm$ standard deviation of the mean squared error ($\downarrow$) after trimming $25\%$ tails for different sparse coding basis functions $M$ and training frames $H$. More basis functions and training samples lead to better reconstruction performance.}
  \label{table_hyperopt}
  \centering
  \begin{tabular}{cccc}
  \toprule
  $M$ $/$ $H$ & $2 \times M$ & $2.5 \times M$ & $3 \times M$ \\
  \midrule
    $512$ & $0.925 \pm 0.129$ & $0.734 \pm 0.125$ & $\mathbf{0.425 \pm 0.106}$ \\
    $1024$ & $0.484 \pm 0.202$  & $0.474 \pm 0.220$ & $\mathbf{0.316 \pm 0.178}$ \\
    $2048$ & $0.195 \pm 0.174$ & $0.309 \pm 0.238$ & $\mathbf{0.190 \pm 0.228}$ \\
  \bottomrule
  \end{tabular}
\end{table}

\begin{figure}[tbp]
\centerline{\includegraphics[width=11cm]{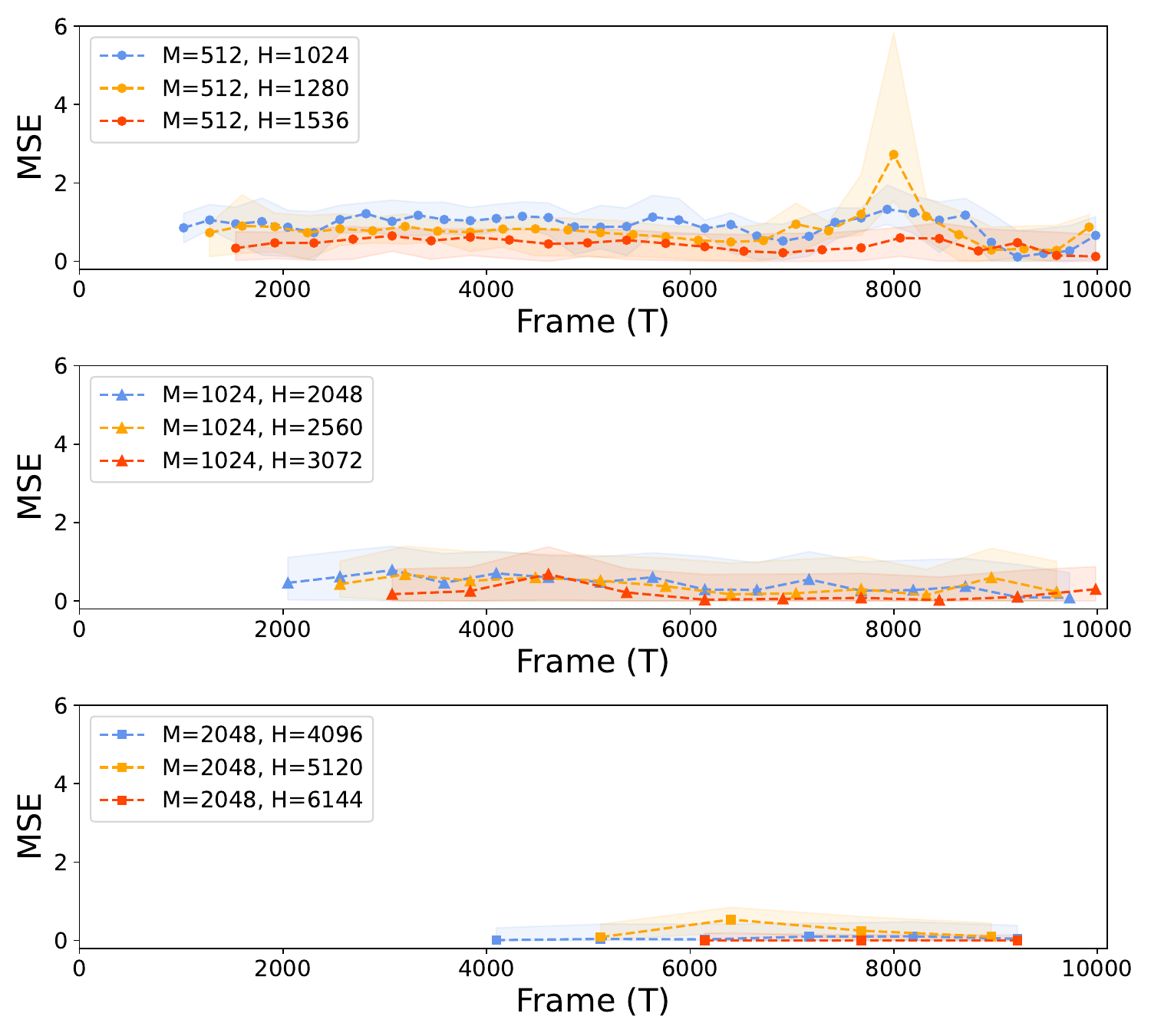}}
\caption{\textbf{Hyperparameter search.} The median and the interquartile range of the mean squared error ($\downarrow$) at each segment for different sparse coding basis functions $M$ and training frames $H$. More basis functions and training samples lead to better reconstruction performance.}
\label{appendix_hyperopt}
\end{figure}

\section{Results}

\textbf{More basis functions and training samples lead to better reconstruction performance.}~
As shown in Table \ref{table_hyperopt} and Figure \ref{appendix_hyperopt}, when using the same number of sparse coding basis functions, the more frames used for training, the lower the average MSEs are generally. 
Additionally, when using more basis functions, the reconstruction performance is better because the learned dictionary is more expressive. 

\begin{figure}[tbp]
\centerline{\includegraphics[trim=20 0 40 0,clip,width=15.5cm]{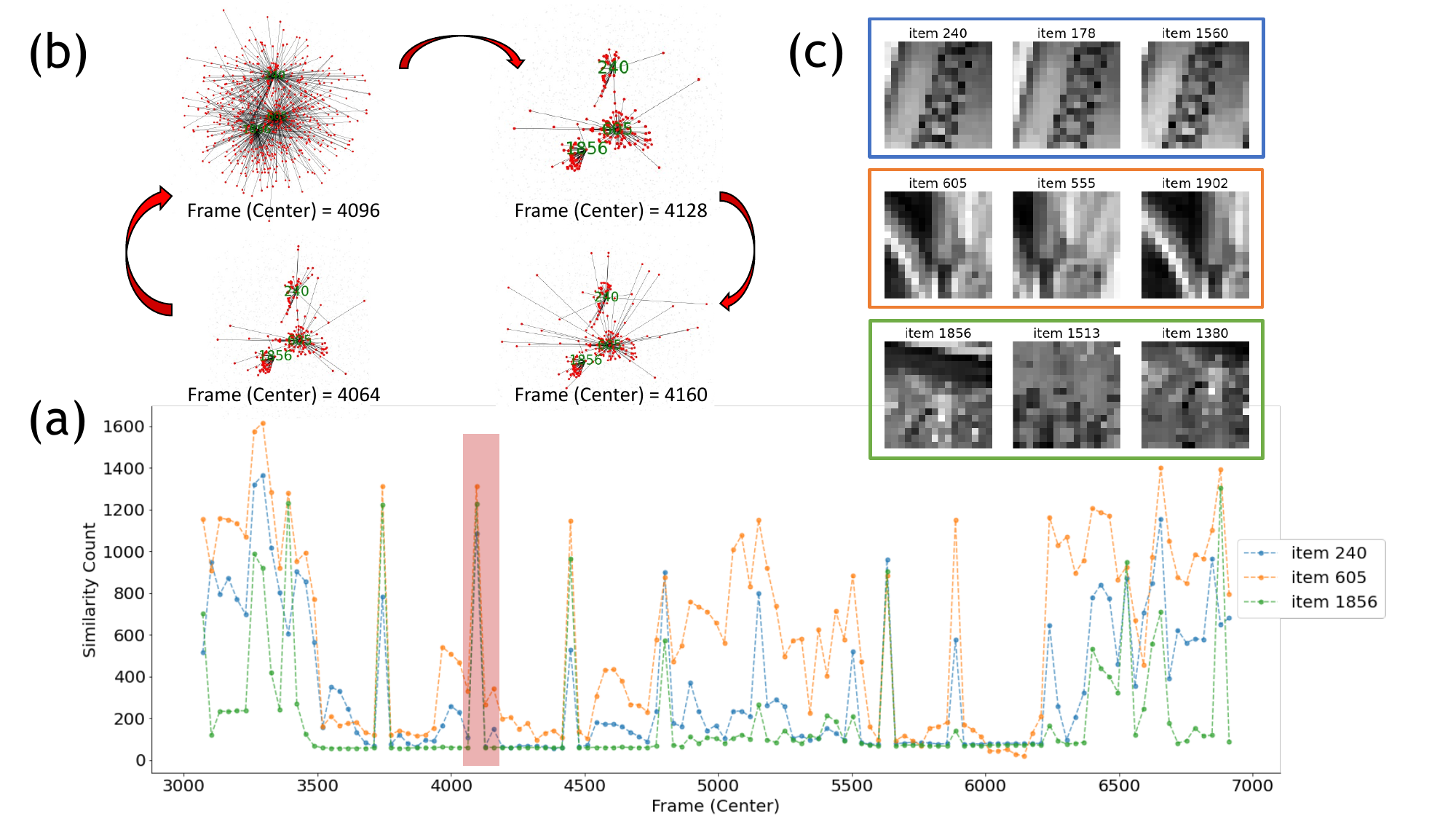}}
\caption{\textbf{Dynamic topological analysis of the PSMT embedding space.} (a) Similarity count of three selected dictionary items (\textcolor{pltBlue}{$240$}, \textcolor{pltOrange}{$605$} and \textcolor{pltGreen}{$1856$}) in graph $\mathcal{G}_t$ constructed from $\mathbf{P^*}_t$ center at frame $t$ using $M=2048$ and $H=6144$ with $\frac{63}{64}$ overlap. Similarity count is computed as the number of dictionary items in the strongly-connected-component containing selected items. (b) Sequence of (shaded in \textcolor{red}{red}) induced subgraph containing selected items. (c) Selected dictionary items with their two most-similar items in $\mathbf{\Phi}_M$.}
\label{fig_dta}
\end{figure}

\noindent\textbf{PSMT embedding space is topologically organized over time.}~
As shown in Figure \ref{fig_dta}, we focus on three dictionary items ($240$, $605$ and $1856$), and each of them is not connected in every $\mathcal{G}_t$ from $\mathbf{P^*}_t$. 
From Figure \ref{fig_dta} (a), we note the actual clusters change over time, which motivates us to iteratively update $\mathbf{P^*}$ for each future frame prediction. 
From Figure \ref{fig_dta} (b), we observe the induced subgraphs from the embedding space $\mathbf{P^*}$ displaying cluster-like behavior and such clusters persist across time. Therefore, each $\mathbf{P^*}_t$ can be topologically organized, so do items in $\mathbf{\Phi}_M$. In Figure \ref{fig_dta} (c), we visualize the corresponding items in $\mathbf{\Phi}_M$ and observe the similarity across the rows and dis-similarity across columns, confirming the topology in $\mathbf{P^*}$ corresponds to that in $\mathbf{\Phi}_M$. 

\begin{figure}[tbp]
\centerline{\includegraphics[trim=0 0 0 0,clip,width=15.5cm]{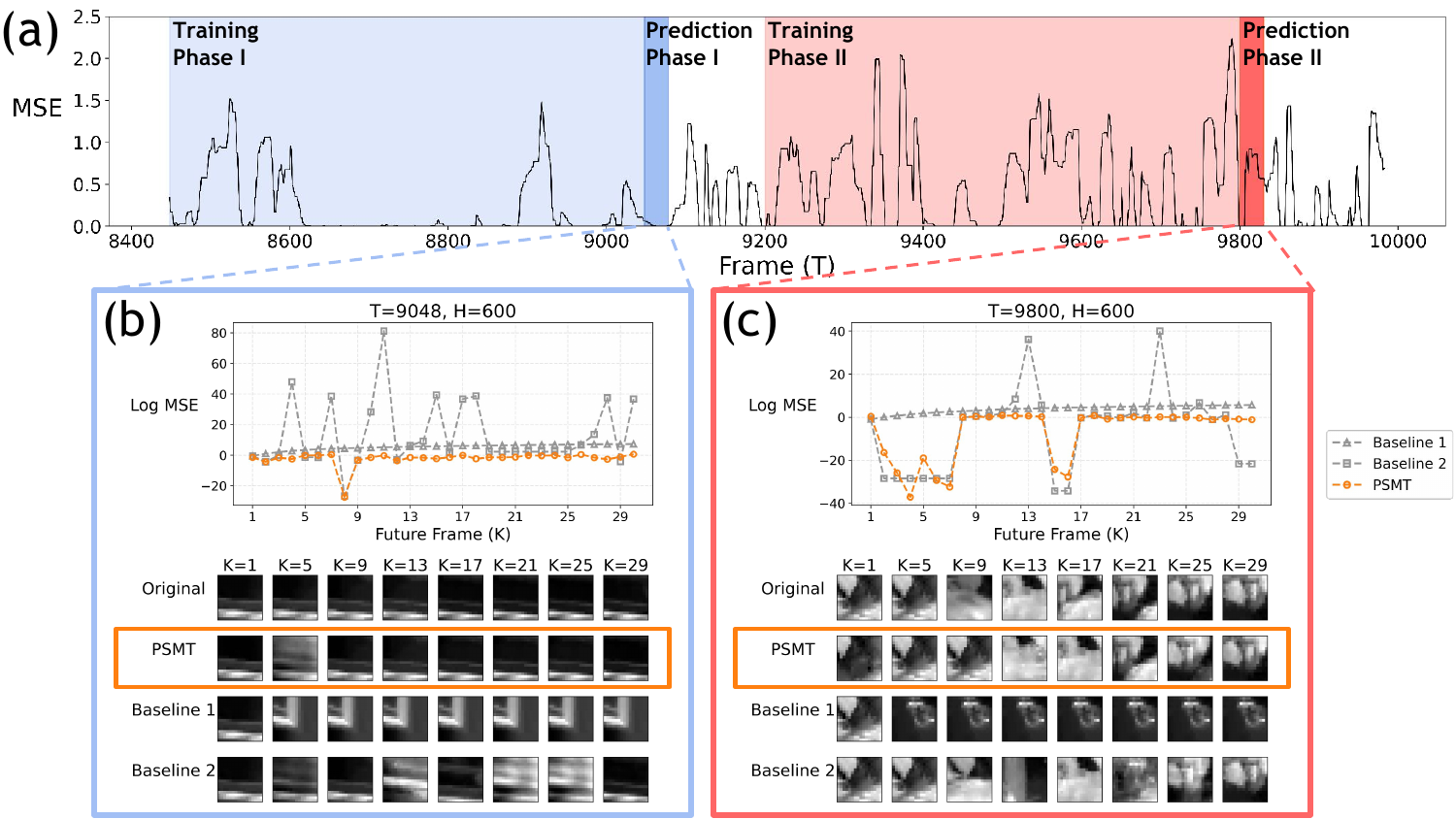}}
\caption{\textbf{Prediction performance.} (a) MSEs between the reconstructed patches and the original ones (frames $8448-9984$) using $512$ basis functions. 
We select two representative phases for prediction, one with relatively static scenes and low MSEs (\textcolor{pltCornflowerBlue}{phase I}) and the other with relatively dynamic scenes and high MSEs (\textcolor{red}{phase II}). 
(b-c) Log MSEs between the predicted and original patches ones (top) and examples of predicted patches (bottom) from prediction phase I ($T=9048$) and II ($T=9800$). 
PSMT outperforms two baselines with lower average log MSEs across $30$ future frames.
}
\label{fig_pred}
\end{figure}

\noindent\textbf{Dynamic PSMT achieves better prediction performance compared to static baselines.}~
We perform prediction on the segment (frames $8448-9984$) with minimum average MSE from $512$ basis functions for the sake of computational efficiency. 
As shown in Figure \ref{fig_pred} (a), we select two representative prediction phases, one with relatively static scenes and low MSEs (phase I) and the other with relatively dynamic scenes and high MSEs (phase II). 
We predict $30$ future frames using PSMT and two other static baseline approaches in these two phases, respectively. 
According to Figures \ref{fig_pred} (b) and (c), PSMT with a dynamic embedding space outperforms two baseline approaches with a static embedding space. 
Specifically, PSMT achieves lower average log MSEs ($-2.18$, $-7.12$) across $30$ future frames in both cases. 
The predicted patches from PSMT better approximate the estimated patches compared to two baselines.

\section{Discussion}
\textbf{Contributions}~
Our contributions in this work are multi-fold. 
First, we propose the PSMT framework for predicting a future temporal sequence of natural scenes. 
Second, we evaluate the PSMT reconstruction performance with respect to two key hyperparameters and show that more basis functions and training samples usually lead to better performance. 
Third, we perform dynamic topological analysis of the geometric embedding matrix over time and demonstrate its dynamic nature. 
Forth, we demonstrate the superiority of PSMT with a dynamic embedding space ($\mathbf{P}^*$ is updated and time-variant) for future temporal sequence prediction compared to two baseline methods with a static embedding space ($\mathbf{P}^*$ is fixed and time-invariant). 

\noindent\textbf{Limitations}~
One important limitation of PSMT is the assumption of temporal linearity (imposed by $\mathbf{D}$) in the geometric embedding space, thus it might not be sufficient to capture sharp or large variations in natural scenes such as a flash of light. 
Another limitation originates from two conditions in sparse coding: 
(1) the number of basis functions is larger than the image dimension such that the learned dictionary $\mathbf{\Phi}_M$ is overcomplete; 
(2) the number of frames is larger than the number of basis functions such that the sparse coefficient matrix $\mathbf{A^*}$ is not singular. 
Thus, small image patches are used to train the model instead of full images. 
Additionally, we use the analytic solution \cite{vladymyrov2013locally} to derive the embedding matrix $\mathbf{P}^*$ but the solution might be too global to capture local dynamics, especially for a large dictionary with many items. Alternatively, stochastic gradient descent with mini-batches can be used to solve for $\mathbf{P}^*$, which better characterizes locality \cite{chen2018sparse}.

\noindent\textbf{Future Work}~
We plan to extend PSMT to learn the dynamic topological organization of brain networks from neuroimaging data, bridging the gap between visual perception and mental representation.

\section{Conclusion}
We present Predictive Sparse Manifold Transform (PSMT), a two-layer unsupervised generative model to learn and predict natural dynamics. 
Our results show that PSMT embedding space is topologically organized over time and PSMT using a dynamic embedding space can achieve better prediction performance compared to static baselines, thus establishing an efficient prediction framework. 

\section{Acknowledgments}
We thank Christopher J. Rozell and Rogers F. Silva for their valuable feedback; Yubei Chen for his help with Sparse Manifold Transform implementation. This work is supported by the Georgia Tech/Emory NIH/NIBIB Training Program in Computational Neural-engineering (T32EB025816).

\clearpage
\bibliography{sample}

\begin{thebibliography}{10}
\providecommand{\natexlab}[1]{#1}
\providecommand{\url}[1]{\texttt{#1}}
\expandafter\ifx\csname urlstyle\endcsname\relax
  \providecommand{\doi}[1]{doi: #1}\else
  \providecommand{\doi}{doi: \begingroup \urlstyle{rm}\Url}\fi

\bibitem[Chen et~al.(2018)Chen, Paiton, and Olshausen]{chen2018sparse}
Yubei Chen, Dylan Paiton, and Bruno Olshausen.
\newblock The sparse manifold transform.
\newblock \emph{Advances in neural information processing systems}, 31, 2018.

\bibitem[Efron et~al.(2004)Efron, Hastie, Johnstone, and
  Tibshirani]{efron2004least}
Bradley Efron, Trevor Hastie, Iain Johnstone, and Robert Tibshirani.
\newblock Least angle regression.
\newblock 2004.

\bibitem[{Gurobi Optimization, LLC}(2023)]{gurobi}
{Gurobi Optimization, LLC}.
\newblock {Gurobi Optimizer Reference Manual}, 2023.
\newblock URL \url{https://www.gurobi.com}.

\bibitem[Hawkins and Blakeslee(2004)]{hawkins2004intelligence}
Jeff Hawkins and Sandra Blakeslee.
\newblock \emph{On intelligence}.
\newblock Macmillan, 2004.

\bibitem[Jiang and Rao(2022)]{jiang2022dynamic}
Linxing~Preston Jiang and Rajesh~PN Rao.
\newblock Dynamic predictive coding: A new model of hierarchical sequence
  learning and prediction in the cortex.
\newblock \emph{bioRxiv}, pages 2022--06, 2022.

\bibitem[Olshausen and Field(1996)]{olshausen1996emergence}
Bruno~A Olshausen and David~J Field.
\newblock Emergence of simple-cell receptive field properties by learning a
  sparse code for natural images.
\newblock \emph{Nature}, 381\penalty0 (6583):\penalty0 607--609, 1996.

\bibitem[Olshausen and Field(1997)]{olshausen1997sparse}
Bruno~A Olshausen and David~J Field.
\newblock Sparse coding with an overcomplete basis set: A strategy employed by
  v1?
\newblock \emph{Vision research}, 37\penalty0 (23):\penalty0 3311--3325, 1997.

\bibitem[Pedregosa et~al.(2011)Pedregosa, Varoquaux, Gramfort, Michel, Thirion,
  Grisel, Blondel, Prettenhofer, Weiss, Dubourg, et~al.]{pedregosa2011scikit}
Fabian Pedregosa, Ga{\"e}l Varoquaux, Alexandre Gramfort, Vincent Michel,
  Bertrand Thirion, Olivier Grisel, Mathieu Blondel, Peter Prettenhofer, Ron
  Weiss, Vincent Dubourg, et~al.
\newblock Scikit-learn: Machine learning in python.
\newblock \emph{the Journal of machine Learning research}, 12:\penalty0
  2825--2830, 2011.

\bibitem[Sokurov()]{RussianArk}
Alexander Sokurov.
\newblock Russian ark.
\newblock URL \url{https://youtu.be/PECz8C7m_Yo}.

\bibitem[Vladymyrov and Carreira-Perpin{\'a}n(2013)]{vladymyrov2013locally}
Max Vladymyrov and Miguel~{\'A} Carreira-Perpin{\'a}n.
\newblock Locally linear landmarks for large-scale manifold learning.
\newblock In \emph{Machine Learning and Knowledge Discovery in Databases:
  European Conference, ECML PKDD 2013, Prague, Czech Republic, September 23-27,
  2013, Proceedings, Part III 13}, pages 256--271. Springer, 2013.

\end{thebibliography}
\newpage
\clearpage
\appendix

\setcounter{figure}{0}
\counterwithin{figure}{section}
\setcounter{table}{0}
\counterwithin{table}{section}

\section{PSMT Algorithm Details}\label{PSMT_detail}
Here, we describe the Predictive Sparse Manifold Transform (PSMT) framework in detail. PSMT is an iterative procedure that uses the past and current image inputs to predict the next input in a sliding fashion. The numerical inputs to PSMT include

\begin{enumerate}[nolistsep]
    \item $T$: the starting point for prediction;
    \item $H$: the window size (i.e., number of most-recent consecutive inputs from the past) used for prediction;
    \item $K$: number of future inputs to be predicted;
    \item $M$: number of items in the overcomplete sparse-coding dictionary;
    \item $N$: number of pixels in the input;
    \item $f$: dimension in the embedded space.
\end{enumerate}

Additionally, we have the past image sequence as $\mathbf{X} = \{\bm{x}_t\}_{t=T-H+1}^{t=T-1}$ and incorporate the future input $\mathbf{X^*}=\{\bm{x}_t\}_{t=T}^{t=T+K-1}$ iteratively in PSMT.

We initialize PSMT by computing overcomplete sparse-coding dictionary (item size $M$) $\mathbf{\Phi}_M \in \mathbb{R}^{N \times M}$ with $N<M$ together with the corresponding non-negative sparse codes $\mathbf{A} = \{ \bm{\alpha}_t \}_{t=T-H+1}^{T-1}$. That is, $\mathbf{A} \in \mathbb{R}^{M \times (H-1)}$ with $M<H$ subject to $\bm{\alpha}_t \succeq 0$, to approximate the input $\mathbf{X}$. Here, the Least Angle Regression (LARS) algorithm \cite{efron2004least} from sklearn \cite{pedregosa2011scikit} is used to solve this estimation: 
\begin{equation}
    \mathbf{X}\approx \mathbf{\Phi}_M \mathbf{A}.
\end{equation}

At each time $t=T+k-1$ for $k=1,...,K$, we receive the input $\bm{x}_{t=T+k-1}$ and want to predict the next input $\hat{\bm{x}}_{t=T+k}$ using PSMT. For the ease of presentation, we use footnotes \textit{prev} to represent $t=T-1+k-1$, \textit{current} to represent $t=T-1+k$, and $\textit{future}$ to represent $t=T+k$.
\begin{enumerate}[nolistsep]
    \item Step 1 to 4 in Algorithm \ref{PSMT_alg}: upon receiving the most recent input $\bm{x}_\text{current}$, we compute its sparse code $\bm{\alpha}_\text{current}$ using the initialized dictionary $\mathbf{\Phi}_M$. We incorporate it into matrix $\mathbf{A}$. Now, $\mathbf{A}\in\mathbb{R}^{M\times (H+k-1)}$. We only include that latest $H$ sparse codes in PSMT by extracting $t=T-H+k-1$ to $t=T-1+k$ from $\mathbf{A}$ into $\mathbf{A^*}$, making $\mathbf{A^*}\in\mathbb{R}^{M\times (H-1)}$:
    \begin{equation}
        \mathbf{A^*} \leftarrow \mathbf{A}_{[t=T-H+k-1: t=T-1+k]}.
    \end{equation}
    \item Step 5 in Algorithm \ref{PSMT_alg}: we solve for the embedding matrix $\mathbf{P^*}$ assuming that the temporal trajectory of $\mathbf{A}$ is linear in the embedding space:
        \begin{equation}
            \mathbf{P^*} := \text{argmin}_{\mathbf{P}} \{ \| \mathbf{P} \mathbf{A^*}\mathbf{D}\|_F^2: \mathbf{P}\mathbf{V}\mathbf{P}^\top=\mathbf{I}\},
        \end{equation}
    where $\mathbf{D} \in \mathbb{R}^{H \times H}$ is a second-order derivative operator where $\mathbf{D}_{t-1,t}=-0.5$, $\mathbf{D}_{t,t}=1$, $\mathbf{D}_{t,t+1}=-0.5$ and other entries are zeros. 
    The closed form solution is $\mathbf{P}^*=\mathbf{U}^\top \mathbf{V}^{-\frac{1}{2}}$ with $\mathbf{P^*}\in\mathbb{R}^{f\times M}$ given by \cite{vladymyrov2013locally} where $\mathbf{V}$ is the covariance matrix of $\mathbf{A^*}$ and $\mathbf{U}$ is a matrix of $f$ trailing eigenvectors of the matrix $\mathbf{Q} = \mathbf{V}^{-\frac{1}{2}} \mathbf{A^*} \mathbf{D} \mathbf{D}^\top \mathbf{A^*}^\top \mathbf{V}^{-\frac{1}{2}}$. 
    \item Step 6 in Algorithm \ref{PSMT_alg}: assume continuous smooth transformation of inputs in the embedding space, adapted from SMT (i.e., $\mathbf{P}^*\bm{\alpha}^*_{t}\approx \frac{1}{2}\mathbf{P}^*\bm{\alpha}_{t-1} + \frac{1}{2}\mathbf{P}^*\bm{\alpha}_{t+1}$), the future frame $t+1$ can be approximated as:
        \begin{equation}
            \mathbf{P}^*\bm{\alpha}^*_{t+1}\approx 2\mathbf{P}^*\bm{\alpha}_t - \mathbf{P}^*\bm{\alpha}_{t-1}.
        \end{equation}
    That is, at $k$, we predict the embedded vector $\bm{\beta^*}$ at $t=T+k$ as
    \begin{equation}
        \bm{\beta}^* := \mathbf{P}^*\bm{\alpha}^*_\text{future}\approx 2\mathbf{P}^*\bm{\alpha}_\text{current} - \mathbf{P}^*\bm{\alpha}_\text{prev}.
    \end{equation}
    \item Step 7 in Algorithm \ref{PSMT_alg}: we recover the future sparse code at $t=T+k$ from $\bm{\beta}^*\in\mathbb{R}^f$ by solving the following optimization using Gurobi \cite{gurobi} with $\mathbf{z} = [ \| \bm{p^*}_1\|_2, \dots,\|\bm{p^*}_N \|_2]^\top$ (i.e., columns in $\mathbf{P^*}$).
    \begin{equation}
        \bm{\alpha}^{\text{REC}}_\text{future} := \text{argmin}_{\bm{\alpha}}  \{ \| \mathbf{P}^*\bm{\alpha}-\bm{\beta}^*\|_F^2 + \lambda \mathbf{z}^\top \bm{\alpha} : \bm{\alpha} \succeq 0\}.
    \end{equation}
    \item Step 8 in Algorithm \ref{PSMT_alg}: we predict the future input at $t=T+k$ by:
    \begin{equation}
        \bm{x}_\text{future} = \mathbf{\Phi}_M \bm{\alpha}^{\text{REC}}_\text{future}.
    \end{equation}
    
\end{enumerate}

\end{document}